\documentclass{article}

\usepackage{PRIMEarxiv}

\usepackage[utf8]{inputenc} % allow utf-8 input
\usepackage[T1]{fontenc}    % use 8-bit T1 fonts
\usepackage{hyperref}       % hyperlinks
\usepackage{url}            % simple URL typesetting
\usepackage{booktabs}       % professional-quality tables
\usepackage{amsfonts}       % blackboard math symbols
\usepackage{nicefrac}       % compact symbols for 1/2, etc.
\usepackage{microtype}      % microtypography
\usepackage{lipsum}
\usepackage{fancyhdr}       % header
\usepackage{graphicx}       % graphics
\usepackage{tabularx}
\usepackage{subfig}
\graphicspath{{media/}}     % organize your images and other figures under media/ folder

\title{Learning Social Cost Functions for \\Human-Aware Path Planning
}

\author{
  Andrea Eirale \\
  Department of Electronics and Telecommunications \\
  Politecnico di Torino \\
  Torino, TO, 10129, Italy\\
  \texttt{andrea.eirale@polito.it} \\
\And
  Matteo Leonetti \\
  Department of Informatics \\
  King’s College London\\
  London, UK\\
  \texttt{matteo.leonetti@kcl.ac.uk} \\
\And
  Marcello Chiaberge \\
  Department of Electronics and Telecommunications \\
  Politecnico di Torino \\
  Torino, TO, 10129, Italy\\
  \texttt{marcello.chiaberge@polito.it} \\
}

\begin{document}
\maketitle

\begin{abstract}
Achieving social acceptance is one of the main goals of Social Robotic Navigation.
Despite this topic has received increasing interest in recent years, most of the research has focused on driving the robotic agent along obstacle-free trajectories, planning around estimates of future human motion to respect personal distances and optimize navigation. However, social interactions in everyday life are also dictated by norms that do not strictly depend on movement, such as when standing at the end of a queue rather than cutting it. In this paper, we propose a novel method to recognize common social scenarios and modify a traditional planner's cost function to adapt to them. This solution enables the robot to carry out different social navigation behaviors that would not arise otherwise, maintaining the robustness of traditional navigation. Our approach allows the robot to learn different social norms with a single learned model, rather than having different modules for each task. As a proof of concept, we consider the tasks of queuing and respect interaction spaces of groups of people talking to one another, but the method can be extended to other human activities that do not involve motion.
\end{abstract}

% keywords can be removed

\section{Introduction} \label{sec:intro}

Robot navigation, especially indoor, has made tremendous progress, enabling the first commercial robots to enter a large number of homes as well as public spaces, such as shopping centers, airports, and restaurants. Most such robots are based on \emph{geometric} navigation, treating people as dynamic obstacles. On the other hand, \emph{social} navigation aims to achieve socially acceptable robot behavior---a central goal of sharing spaces with service robots.

Most of the work in social navigation is focused on the direction and velocity of the people surrounding the robot, to respect personal space and act naturally in scenarios that require coordination, such as narrow corridors. Our work is orthogonal to such efforts and considers how robot path planning should be affected not by people's velocity but by their activity. Indeed, even when people do not move at all, such as in a queue, the robot should not just consider them as static obstacles and cut in front of them. 

We propose a learning-based approach that leverages existing widespread planners. We focus the learning effort on the social aspect, rather than on areas of navigation in which current planners are very effective, such as obstacle avoidance. For this reason, contrary to end-to-end learning, we propose to learn an additional term of the cost function to use in planning, while retaining the default behavior everywhere else. Planners, to be efficient, work on a cost map, while social navigation depends on many other aspects, such as whether two obstacles are, in fact, people, and whether they are talking to each other or not. Our model takes in input a representation of the position of the people in the vicinity of the robot and the position of the goal, and outputs a social cost map in addition to the usual obstacle one. We present an architecture and a methodology to train such a neural network, which effectively projects the social navigation problem into classical 2D navigation.

As a proof of concept, we use two common situations that appear frequently in public spaces: people queuing and people talking in small groups. Our methodology is not tied to these particular scenarios and can be extended to any other setting in which practitioners can provide a dataset. Both scenarios have been extensively studied separately. We show how the same method can effectively plan for both scenarios at the same time, with the potential to incorporate many more.
We evaluate the generalization abilities of the proposed architecture; we analyze the generated plans in a simulated environment in Gazebo~\cite{koenig2004design}, and we demonstrate the resulting behaviors in the real world on a Pal Robotics TIAGo.  

\section{Related Work} \label{sec:related}

The first examples of designing a behavioral policy able to consider social factors date back more than twenty years, with the deployment of RHINO \cite{burgard1999museum} and MINERVA \cite{thrun2000probabilistic} as tour guides in museums. In those cases, the robotic system perceived people as dynamic, non-responsive obstacles. In the following years, researchers focused on systems able to distinguish humans from inanimate objects. 

\subsection{Motion-based Human-Aware Navigation}
Many works focused on proxemics, inflating and reshaping the cost function of people in the scenery, depending on the velocity and the facing direction of each person \cite{kirby2010social, kollmitz2015time, scandolo2011anthropomorphic, fang2020human}. One of the most well-known solutions in this category is the Robot Operating System (ROS) social navigation layers~\footnote{\url{http://wiki.ros.org/social_navigation_layers}}, which alter the cost map with a Gaussian distribution around people, with an increased cost in the direction of motion. Mateus et al. \cite{mateus2019efficient} exploit deep learning with hybrid asymmetric Gaussian functions to represent proxemics for modeling comfort distance with pedestrians. Truong and Ngo \cite{truong2016dynamic} took states (position, orientation, motion, and hand poses) and social interaction information relative to the robot into account to model extended personal space and social interaction space using two-dimensional Gaussian functions. 

Social Force Model (SFM) \cite{helbing1995social} is another popular technique to model human and robot motions. Ferrer et al. \cite{ferrer2014proactive} extend the SFM to present a proactive navigation approach, where the robot can produce actions with a minimum impact on surrounding pedestrians. Yang et al. \cite{yang2019socially} employ the SFM on an omnidirectional robot to achieve natural and human-like movements.
More recent works consider autonomous agents able to plan an optimal trajectory, avoiding collisions with people and obstacles, by planning around estimates of future human motion. These predictions are often achieved with deep neural networks, like generative adversarial networks \cite{gupta2018social}, convolutional neural networks \cite{mohamed2020social, zhao2020noticing}, and attention transformers \cite{vemula2018social}.

Intention-aware navigation is the direct evolution of the previous methods. In this case, by exploiting behavior prediction models, it is possible not only to predict the future movements of humans but also to estimate their final goals. Bai et al. \cite{bai2015intention} model the uncertainty of human intent in the Partially Observable Markov Decision Process (POMDP) framework. Navigation in dense human crowds is another interesting task, where solutions focused on avoiding blockage due to human activities \cite{park2016hi} and cooperating with people through interacting Gaussian processes \cite{trautman2015robot}. Mavrogiannis et al. employ geometric \cite{mavrogiannis2019multi} and topological invariant \cite{mavrogiannis2020multi} representations to model the coupling among trajectories of multiple navigating agents.

Deep Reinforcement Learning has been used in many works for prediction in crowd navigation domains. Chen et al. \cite{chen2017decentralized} apply CADRL, a deep reinforcement learning framework for socially aware multi-agent collision avoidance. Everett et al. \cite{everett2018motion} exploit an actor-critic variant to relax prior assumptions and learn policies and agent motion models simultaneously. Furthermore, Tai et al. \cite{tai2018socially} train a generative adversarial imitation learning model on a dataset generated using the social force model \cite{helbing1995social}. Finally, Chen et al. \cite{chen2019crowd} use attention-based reinforcement learning to produce interaction-aware collision avoidance behaviors. 

The works cited above, as well as the majority of scientific literature on social navigation, focus on people's movements: their trajectory, intention, and destination. However, several social norms are independent of movement but dependent on people's activity. For instance, walking between two people talking to each other is considered rude, even if they are just standing still. Such scenarios are currently under-represented in the scientific literature and form the basis of this work.

\subsection{Task-specific Social Navigation}
Other works have been more focused on solving specific social tasks during navigation. Xiao et al. \cite{pmlr-v205-xiao23a} train a Performer architecture \cite{choromanski2020rethinking} in an imitation-learning fashion to learn a cost function and improve an MPC controller. This system is then used to solve different social tasks, such as moving around corners and respecting comfort distance in a human-aware manner. However, unlike our approach, Performer-MPC requires the cost function to be learned individually for each navigation scenario.

Within the queue following, one of the social navigation scenarios we consider in this paper, Nakauchi et al. \cite{nakauchi2002social} designed a dedicated pipeline. Their work focuses on how a line of people can be defined and then perceived by the autonomous agent. The navigation system generates a series of goals (depending on the number of people in the queue) until the back of the line is reached, and does not provide full path planning. In \cite{banisetty2021socially, 9515424}, Banisetty et al. exploit simple geometric reasoning to compute a social goal at the beginning of the line. The true goal of the navigation, placed at the end of the line, is then replaced with the social goal. However, this system only allows the robot to join the queue and is unable to follow the line until the robot reaches the final goal.

Many works focused on detecting groups of people, based on the definition of F-formations provided by Kendon \cite{kendon1990conducting}. Cristiani et al. \cite{cristani2011social} and Setti et al. \cite{setti2015f} proposed Hough Voting with the people's positions and head orientations to detect F-formations, while Hedayati et al. \cite{hedayati2020reform, hedayati2019recognizing} treated the problem as a binary classification. A few works focused on unsupervised group detection in dynamic, egocentric views \cite{taylor2020robot, schmuck2020rica}, while Schmuck et al. \cite{schmuck2021growl, schmuck2022igrowl} proposed GROup detection With Link prediction (GROWL), demonstrating the effectiveness of Graph Neural Network in the detection of people interaction.
While our work does not strictly focus on recognizing the complex connections within groups of people, our experimentation shows that, for social navigation purposes, implicit knowledge of the interaction between people arising from simple spatial information is often sufficient to recognize and avoid sparse groups of people. However, group detection could be integrated within our methodology to further improve the navigation, especially through dense crowds.

Differently from all the work described above, our approach is targeted at all scenarios in which people are static, and yet their position or activity dictates socially acceptable robot paths. 

\section{Background}

Grid-based path planners, such as Dijkstra's or A*, work with a discrete representation of the environment, which can be formalized as a two-dimensional, undirected graph $M=(V, E)$, usually referred to as a \emph{grid map}. In this graph, $V$ is a vertex set of size $n\times m$, where each vertex corresponds to a cell in the map, and $E$ is an edge set of size $k$. Edges connect adjacent map cells. A path is a sequence of vertices $P=(v_1, v_2,...,v_n)$ such that there exists an edge $e_{i,i+1}\in E$ connecting $v_i$ and $v_{i+1}$, with an associated distance $d_{i,i+1}$ for $1\leq i<n-1$. 

Given a starting vertex $v_s\in V$ and a goal $v_g\in V$, a shortest feasible path from $v_s$ to $v_g$ is defined as $P^*=(v_1, v_2,...,v_n)$, where $v_1=v_s$ and $v_n=v_g$. This path is a minimizer of:
\begin{equation}
\label{eq:shortest_path}
    P^*=arg\,min_{v1, ..., vn \in V}\sum_{i=1}^{n-1}{f_c(v_i, v_{i+1})},
\end{equation}
where $f_c:V \times V \rightarrow \mathbb{R^+}$ is a \textit{cost function}.

Cost functions can be defined with several terms, encoding both the distance between two nodes and the desirability of each node. A \emph{cost map} $C$ corresponding to a grid map $M$ is an $n \times m$ matrix, with a cost $c_i \in \mathcal{R^+}$ for each element $v_i \in V$, representing the cost of entering the node.  Usually, a priori information about the environment, like walls and other static obstacles, is gathered in a global cost map $C_G$. Sensor data, containing the detection of obstacles around the agent, are instead included in a local cost map $C_L$ centered on the robot's reference frame. The total cost of a transition is, therefore: 
\begin{equation}
\label{eq:planner_cost_function}
    f_c(v_i, v_{i+1}) = d_{i,i+1} + c_{G,i+1} + c_{L,i+1} \; ,
\end{equation}
where $c_{G,i+1}$ and $c_{L,i+1}$ are the costs of $v_{i+1}$ respectively in $C_G$ and $C_L$.

\begin{figure}[ht]
    \centering
    \includegraphics[width=1.0\columnwidth]{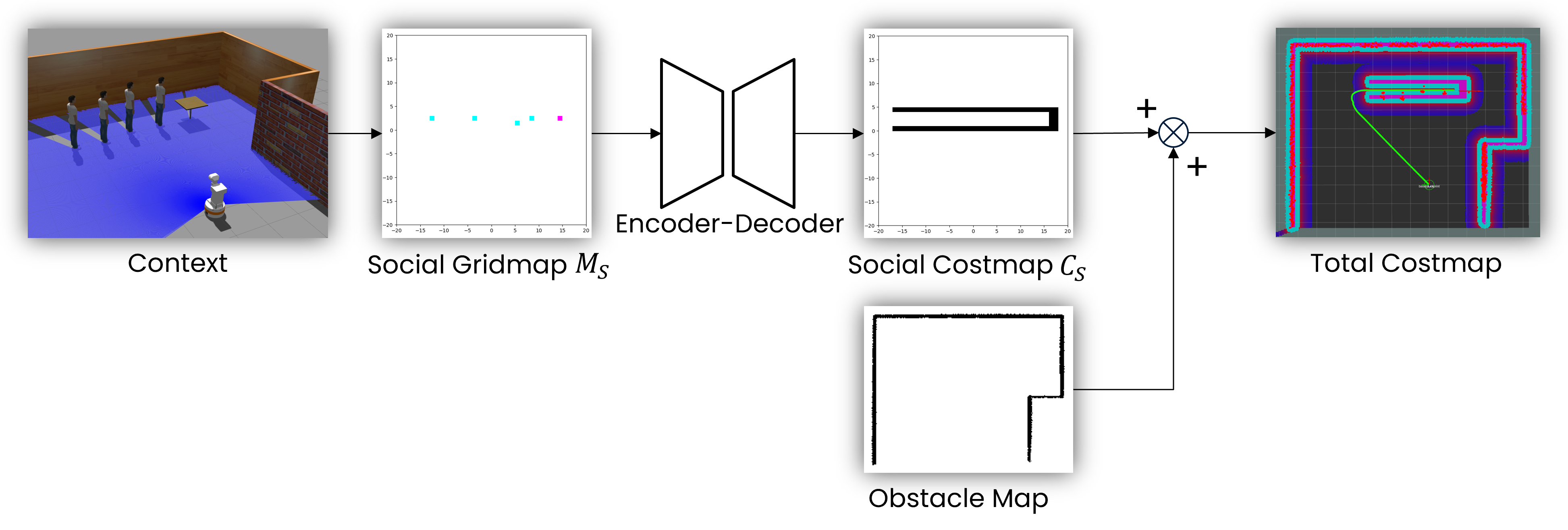} 
    \caption{Representation of the pipeline used to integrate the social costs into the traditional navigation system cost function. The social grid map $M_S$ is obtained from the position of the goal and people. The planner will compute the shortest path towards the goal considering the social cost map $C_S$ in addition to all the other obstacle cost maps.}
    \label{fig:social_costmap_diagram}
\end{figure}

\section{Methodology} 

We want to retain all the desirable properties of grid-based navigation systems, such as path optimality and obstacle avoidance, while introducing socially acceptable behaviors. All social aspects that require the robot to avoid a particular node in the map can be encoded in an additional cost map. We learn the social cost function $f_S : M_S \rightarrow C_S$ through a deep neural network, able to generate the social cost map $C_S$ encoding \emph{social obstacles}. 

Figure \ref{fig:social_costmap_diagram} shows the pipeline of the methodology. Positions of the goal and people are gathered into a social local grid map $M_S$, centered on the robot's reference frame. A learned encoder-decoder produces the social cost map, which is added to the other navigation costs to produce the total cost map. This process repeats continually, as the social cost map is regularly updated to reflect the latest position of the people in the scene.

The total cost expressed in Equation \ref{eq:planner_cost_function} can then be rewritten as:
\begin{equation}
\label{eq:proposed_formulation}
    f_c(v_i, v_{i+1}) = d_{i,i+1} + c_{G,i+1} + c_{L,i+1} + c_{S,i+1} \; ,
\end{equation}
where $c_{S,i+1}$ is the costs added by the social cost map $C_S$.

In the rest of this section, we will describe the network's architecture of the social cost function $f_S$, how the training dataset is created, and how the network is deployed in a real system.

\subsection{Network Architecture}
\begin{figure}[ht]
    \centering
    \includegraphics[width=1.\columnwidth]{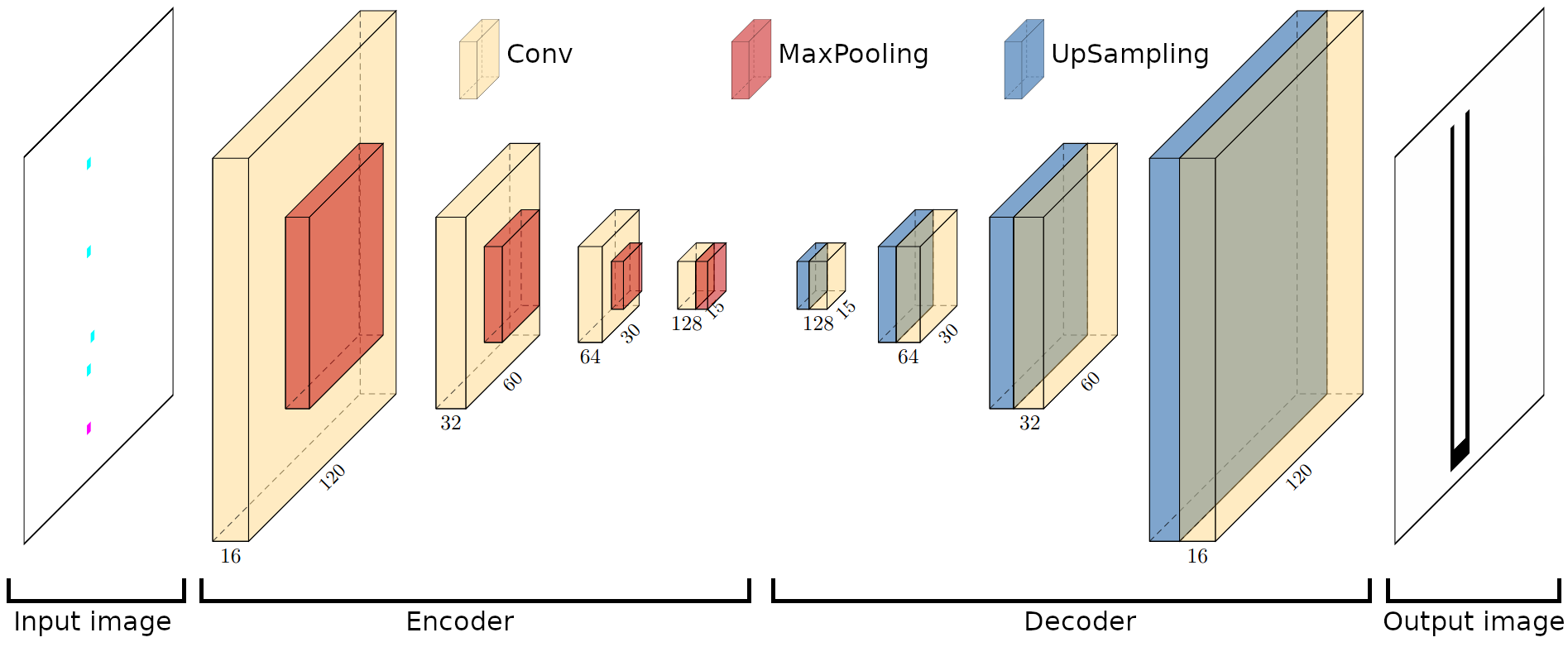} 
    \caption{Representation of the Encoder-Decoder neural network used in this work.}
    \label{fig:Network_structure}
\end{figure}
The network is an encoder-decoder consisting of convolutional, maxpooling, and upscaling layers with $342,049$ (learnable) parameters. The network structure is shown in Figure \ref{fig:Network_structure}. The input discretizes a 24x24m area with a resolution of 0.2m, represented in a grid map with 120x120 cells. These measures are compatible with a typical indoor Lidar sensor, usually in the 8--18m range. 

\subsection{Dataset creation}

We generate the dataset with a series of RGB images for each human activity we want the system to react to. The dataset consists of pairs $\{(M_S^{(i)}$, $C_S^{(i)})\}$  of grid map and corresponding cost map.

The maps for a particular human activity only have elements pertaining to that activity, excluding other elements, such as obstacles. The label image, that is, the intended output of the network, is a grey-scale cost map with the same dimensions as the input image, representing only the costs corresponding to the input map. 

As a proof of concept, in this work, we consider the social scenarios of a queue of people and that of small groups of people talking. In principle, any scenario that only requires the robot to avoid certain areas can be added.

\subsubsection{Queuing}
In this scenario, we expect the robot to follow the queue to the goal without cutting the line. We create grid maps with a line of people behind a goal, represented with two different colors, so that the network can distinguish the people from the goal. Each sample we generate is randomized through a series of varying parameters: the number of people in the queue (which also influences the total length of the queue), the distance between two people, the distance between the first person and the goal, and the deviation of each person from the center axis of the queue. The label image represents a U-shaped obstacle surrounding the queue. The thickness of the U-shaped obstacle walls should be tuned according to the robot's footprint, in order for the planner to plan around the queue and leave the only available path toward the goal behind the queue (cf. Figure \ref{fig:social_costmap_diagram}). This ``social corridor" towards the goal should be narrow enough to force the robot to wait behind the last person.  

\subsubsection{Groups of People}
In this scenario, we expect the robot to go around groups of people rather than through them. The input grid map contains the small groups of people. In this case, the randomized parameters are: the number of people for each group, their position in the map, and the number of groups. The label image consists of a variable number of virtual obstacles (depending on the position of the people), filling the gap between them.

\subsubsection{Further refinements}
Each training image contains at least one instance for each considered social scenario, positioned randomly within the grid map space. To improve generalization, we also diversify the dataset according to the following criteria. In the queuing scenario, with a small probability, the training image does not include the queue. Instead, it has the goal in a random position, with the label image not containing any virtual obstacle. This variation prevents the network from associating the social obstacle with the goal alone. Similarly, to prevent the network from associating the obstacle with each person instead of a group of people, a certain number of isolated people have a probability of spawning in the image. Each generated sample pair is oriented differently to obtain rotation invariance in the prediction.

Every sample pair of grid map and label is created programmatically with an automatic generator, which may raise the question: Can such a generator be used online---do we need the neural network at all? The generator for each scenario can produce pairs $(M_S^{(i)}, C_S^{(i)})$, but it is not equivalent to the function $f_{S}$ that translates a general social grid map into a cost map. The generator first creates the social obstacles in randomized locations and then places people and the goal. This is easier than the inverse operation, and works on a single scenario. For this reason, generators are used only to produce samples for the function $f_{S}$, which is then learned to classify a given grid map into its corresponding cost map and to do so with possibly different scenarios simultaneously.

\subsection{Deployment}

We deployed the method in simulation and on a real robot by integrating it into a ROS2 \cite{ros2} node. Periodically, this node retrieves the position of the goal and people and converts this information into the input grid map image representation $M_S$ used by the network. The generated image is then provided to the neural network, and the output prediction of the social cost map $C_S$ is published on an OccupancyGrid ROS2 topic. This map is used as a local cost map by the Nav2 \cite{macenski2020marathon2} planner, in addition to all the other traditional cost maps provided to the navigation system. From Nav2 parameters, we select a wavefront Dijkstra expanded holonomic planner in combination with a Model Predictive Path Integral Controller (MPPI). In order to wait in line behind each person, we use a simple Nav2 behavior tree with \textit{Wait} nodes in combination with a short obstacle maximum detection range, which enables the robot to reach the queue and then wait for the following person to proceed, freeing the computed path, before advancing. 

\section{Experimental Validation} \label{sec:results}

In this section, we first evaluate the generalization of the learned social cost function, and then demonstrate the resulting behaviors in a number of scenarios both in simulation and on a real robot. Our focus is on evaluating the system's ability to accurately recognize and handle the predefined static social tasks. Other social factors, such as proxemics, which involve understanding and maintaining appropriate personal space, or navigating dynamic and crowded environments, are not taken into account. Our method is designed as a modular plug-in that can be seamlessly integrated with other planners based on cost maps, which can address any additional aspects of social navigation.

\subsection{Learning Evaluation}
The network was trained with a generated dataset containing $5\times10^5$ samples for 5 epochs on a machine with an Intel Core i7-9700K CPU and an Nvidia GeForce RTX 2080 Ti, taking approximately 30 minutes. Although the generated cost maps may show minor discrepancies from the expected correct labels in a few cells, they may still be capable of producing socially acceptable paths. That is, what would increase the error in terms of the accuracy of cost map generation, actually results in a perfectly acceptable behaviour. For this reason, we measure the success rate of the generated cost maps rather than relying solely on the accuracy of the cost map on the test set. A cost map is deemed successful if the path it generates adheres to social norms---specifically, if it follows the queue correctly and avoids navigating through or disrupting groups. Conversely, a cost map is considered unsuccessful if the introduced costs lead to navigation failures or result in paths that violate social norms, such as cutting across groups or deviating from the queue.

We produced three different test sets, containing respectively images only with queues, only with groups, and with queues and groups mixed. Table \ref{tab:results} shows the parameters for training, testing, and success rates. In every dataset we generated for training and testing, where the queue appears, the distance between two people in the line and from the first person and the goal is in the range 0.5-1.5 meters, while the deviation of each person from the center axis of the queue is in the range 0-0.5 meters from each side~\footnote{The source code we used to generate each dataset and to train the network is available at \url{https://github.com/PIC4SeR/SocialCostFunction}}. In all cases the success rate is above $95\%$, which we consider sufficiently reliable.

\begin{table}[]
\centering
\caption{Parameters used for training and test datasets. The "Success" column represents the percentage of samples in which the network could correctly recognize and delimit each social scenario.}
\label{tab:results}
\renewcommand{\arraystretch}{1.5}
\begin{tabularx}{\textwidth} { 
    >{\raggedright\arraybackslash}X 
    >{\centering\arraybackslash}X 
    >{\centering\arraybackslash}X 
    >{\centering\arraybackslash}X 
    >{\centering\arraybackslash}X 
    >{\centering\arraybackslash}X }
    \hline
    \textbf{Dataset} & \textbf{Samples} & \textbf{People in queue} & \textbf{People in group} & \textbf{Number of groups} & {\textbf{Success}}\\
    \hline
    \textbf{Train} & 5e5 & 2 to 5 & 2 & 1 & n/a \\
    \hline
    \textbf{Test queue only} & 400 & 2 to 15 & n/a & n/a & 95.75\% \\
    \hline
    \textbf{Test groups only} & 300 & n/a & 2 to 7 & 1 to 5 & 98.67\% \\
    \hline
    \textbf{Test queue and groups} & 300 & 2 to 9 & 2 to 4 & 1 to 3 & 97.33\% \\
    
    \hline
\end{tabularx}
\end{table}

\begin{figure}[ht]
    \centering
    
    \subfloat[][\emph{Gazebo simulation environment}]
        {\includegraphics[width=52mm]{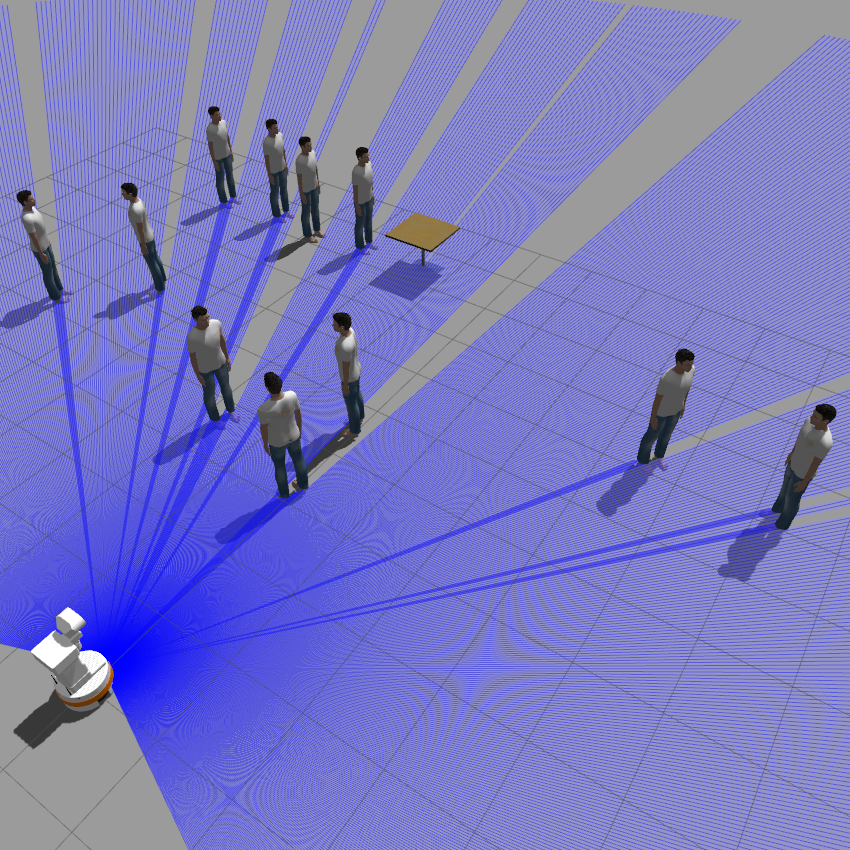}} \quad
    \subfloat[][\emph{With Social cost map}]
        {\includegraphics[width=52mm]{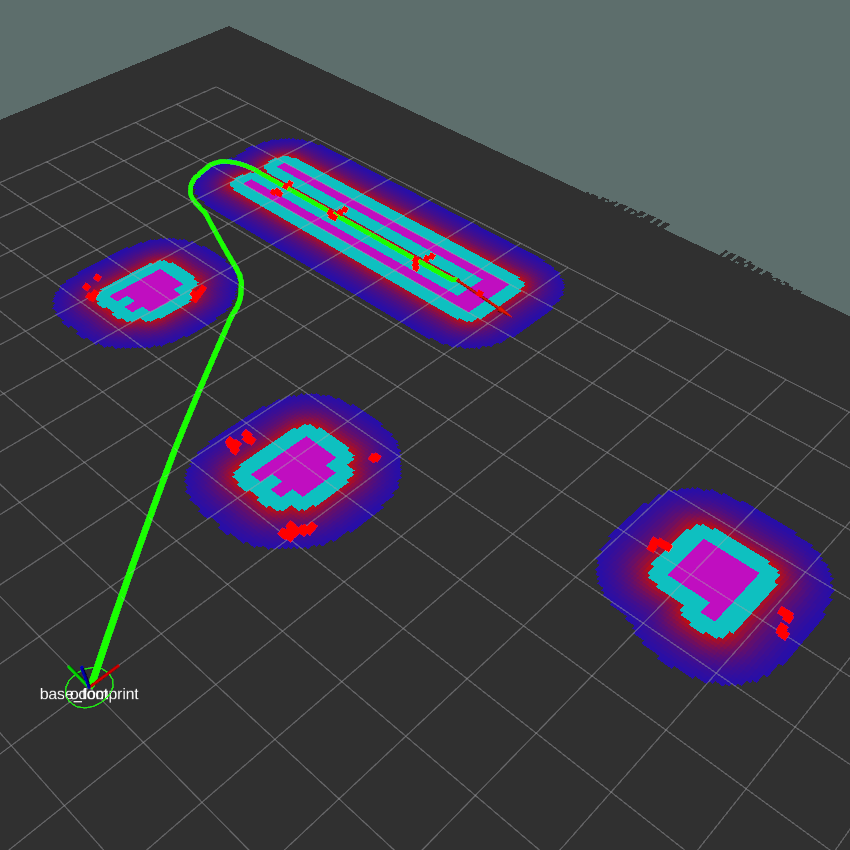}} \quad
    \subfloat[][\emph{Without Social cost map}]
        {\includegraphics[width=52mm]{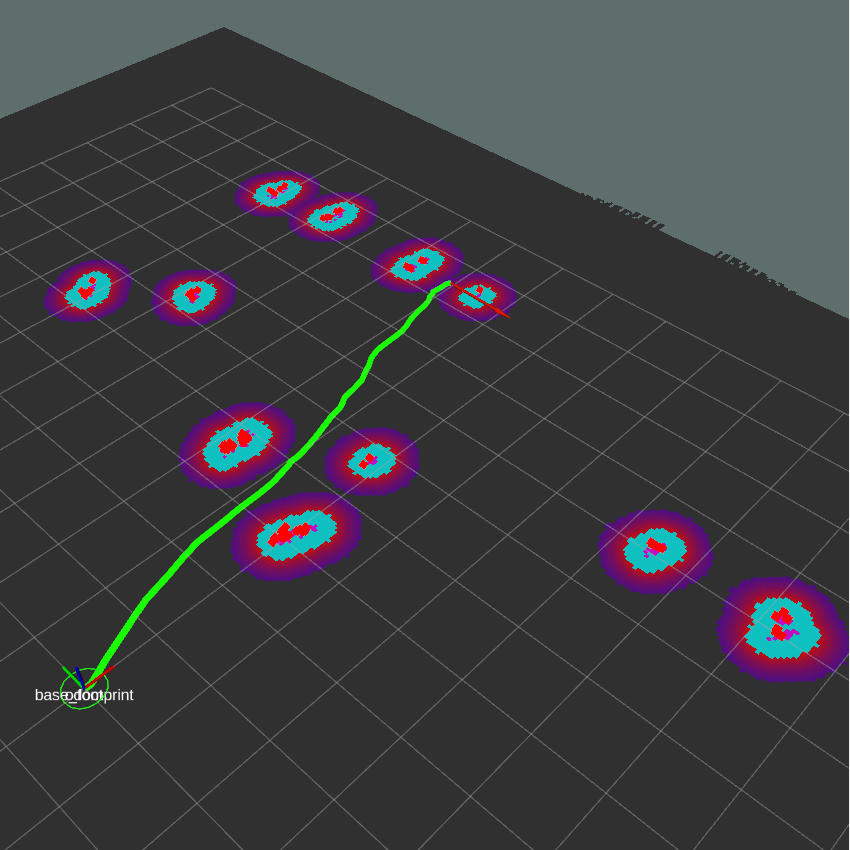}} \\
        
    \caption{Simulation of navigation in a demo environment with three different groups of people interacting with each other and one queue, visualized in $(a)$ with Gazebo. Magenta patches in the cost map represent areas with high costs corresponding to obstacles, while the cyan-to-blue gradient is a decreasing cost that inflates the obstacles. The green line is the computed path towards the goal, positioned at the end of the queue. In $(b)$, the social cost map is included, allowing the navigation algorithm to plan a path that drives the robot at the beginning of the queue without crossing any group of people. On the contrary, in $(c)$ the social cost map is not considered; the robot cuts the line directly to the goal and passes between a group of people talking.}
    \label{fig:general_sim}
\end{figure}

\begin{figure}[ht]
    \centering
    
    \subfloat[][\emph{Front}]
        {\includegraphics[width=52mm]{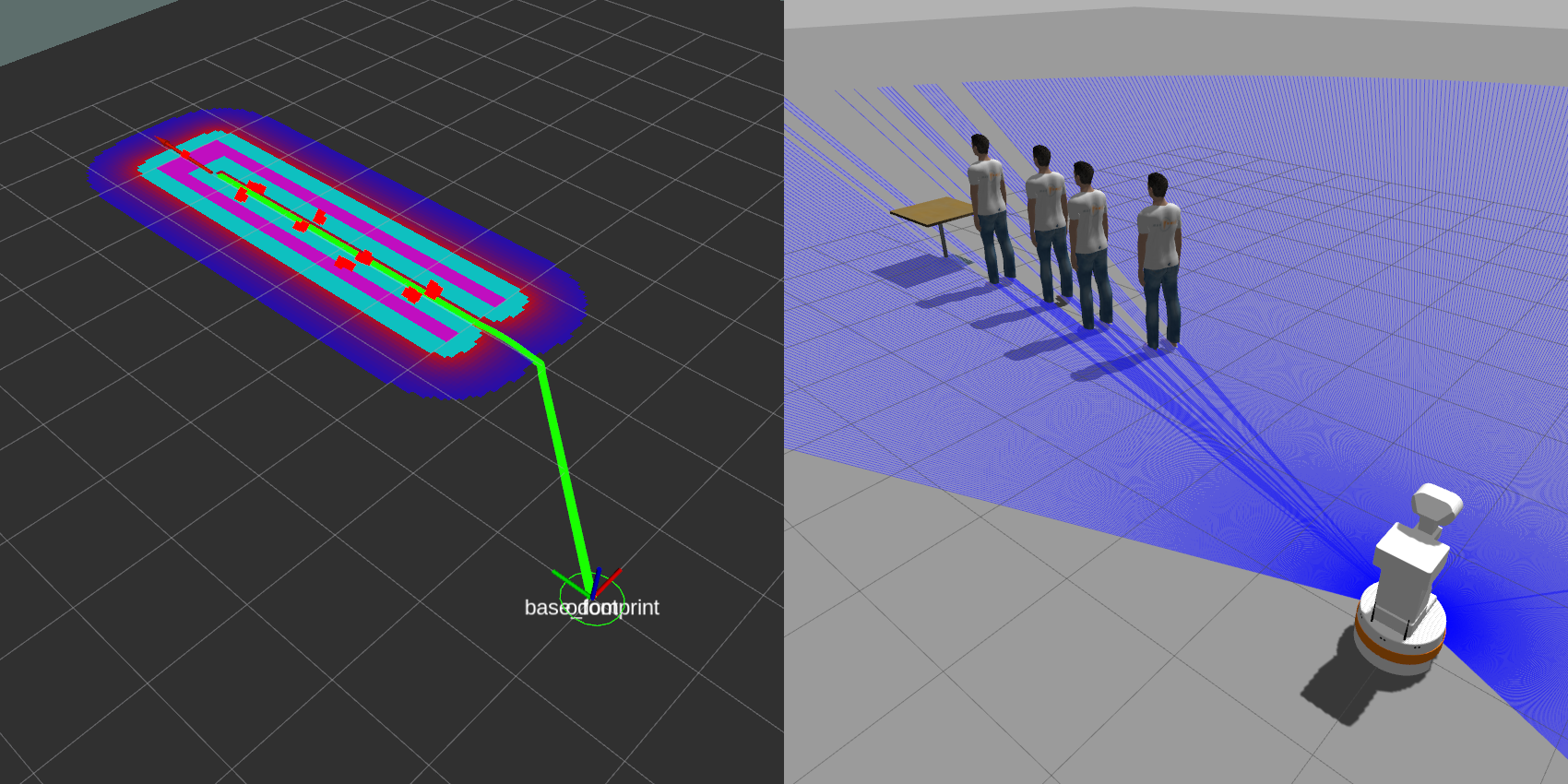}} \quad
    \subfloat[][\emph{Side}]
        {\includegraphics[width=52mm]{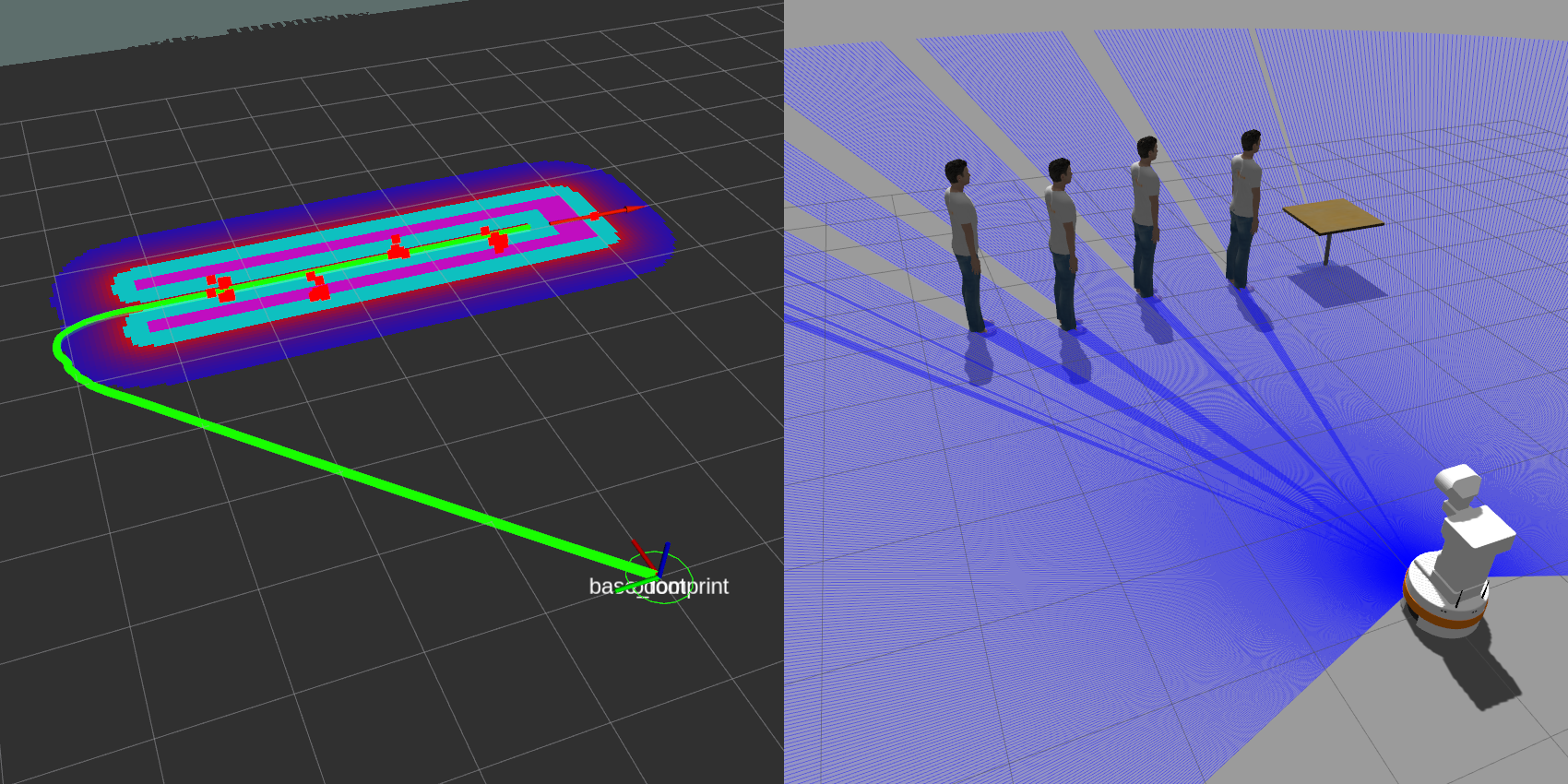}} \quad
    \subfloat[][\emph{Back}]
        {\includegraphics[width=52mm]{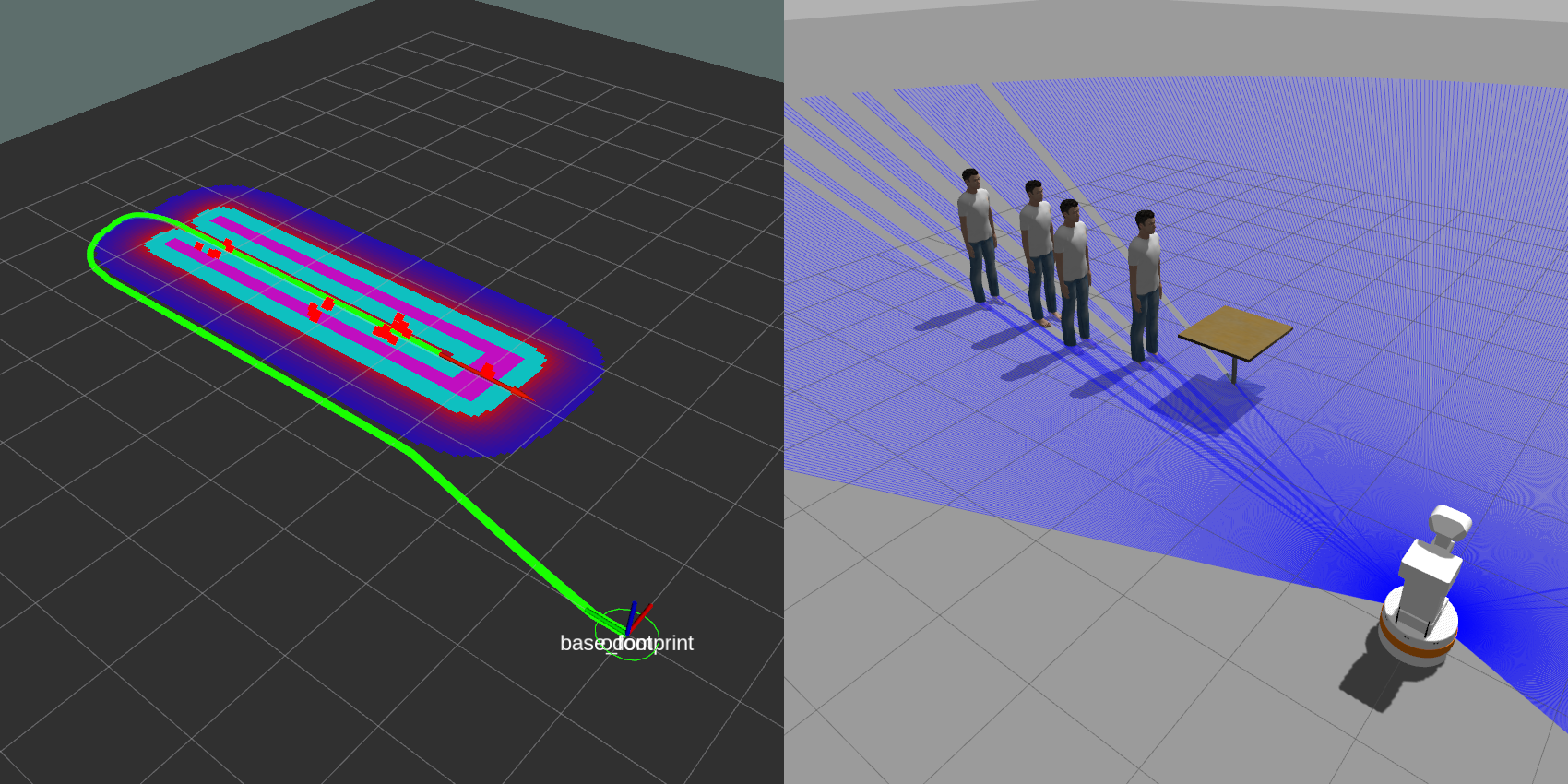}} \\
        
    \caption{Simulation of the robot approaching the queue from different directions: from the front $(a)$, the side $(b)$, and the back $(c)$ of the line of people. As can be seen, with the social cost map, the planner is always able to set the robot in the queue correctly.}
    \label{fig:approaching_direction}
\end{figure}

\begin{figure}[ht]
    \centering
    
    \subfloat[][\emph{}]
        {\includegraphics[width=80mm]{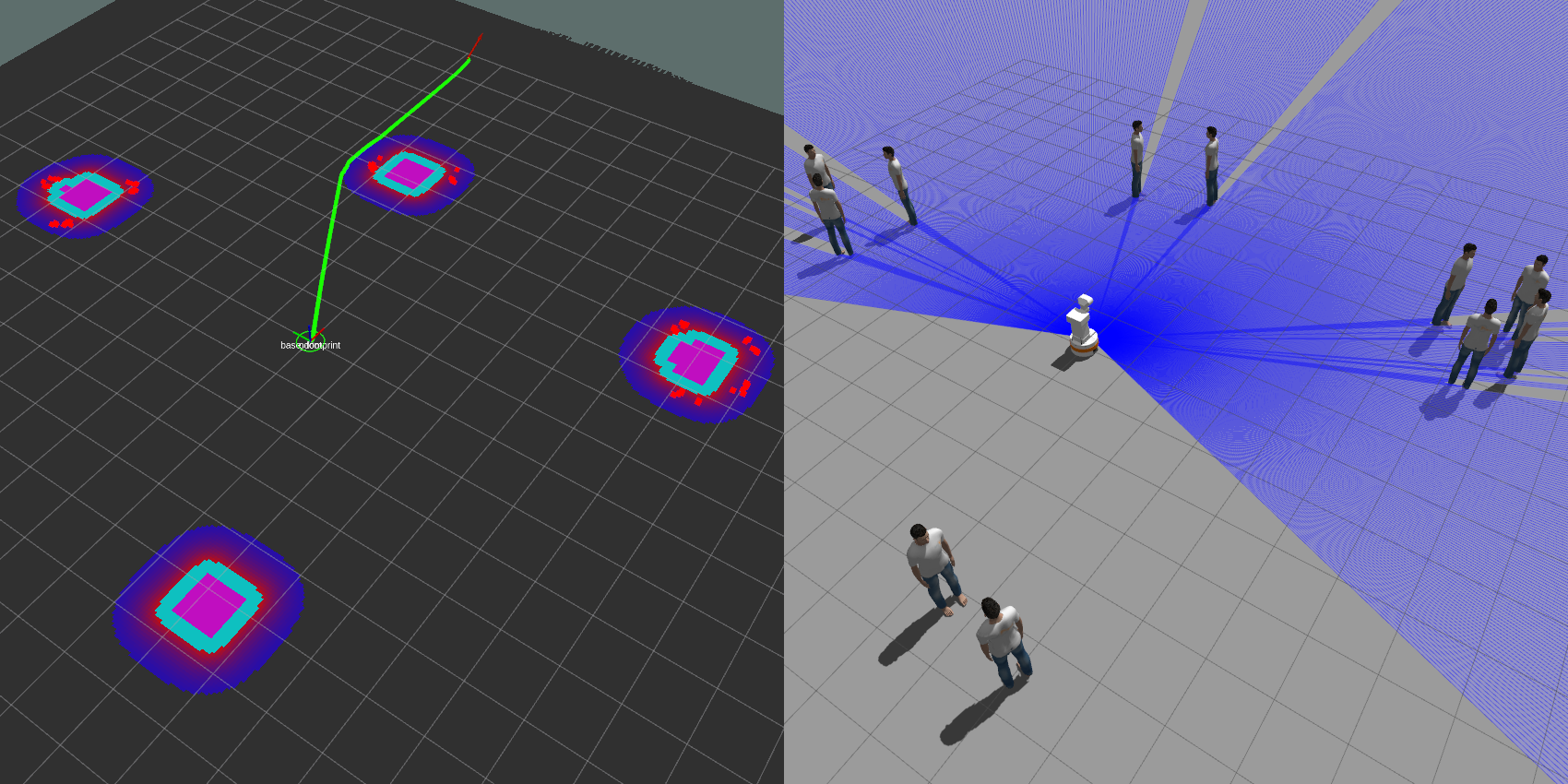}} \quad
    \subfloat[][\emph{}]
        {\includegraphics[width=80mm]{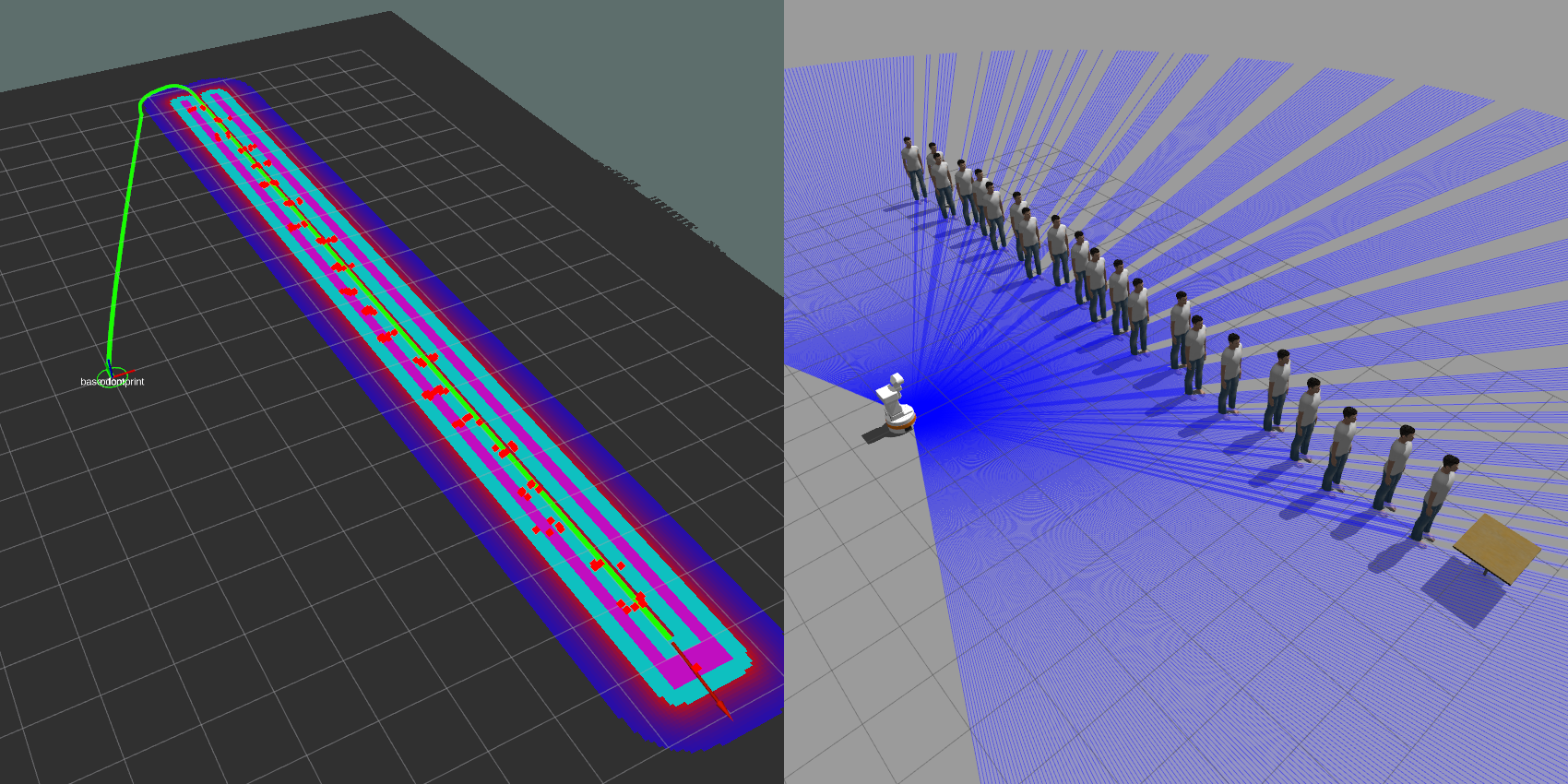}} \\
        
    \caption{Simulation of the robot moving between multiple groups of people $(a)$ and approaching a long queue of people $(b)$. The network can recognize and delimit the whole queue and each group, assigning high costs to the interaction areas between the people.}
    \label{fig:long_queue}
\end{figure}

\begin{figure}[ht]
    \centering
    \includegraphics[width=0.8\columnwidth]{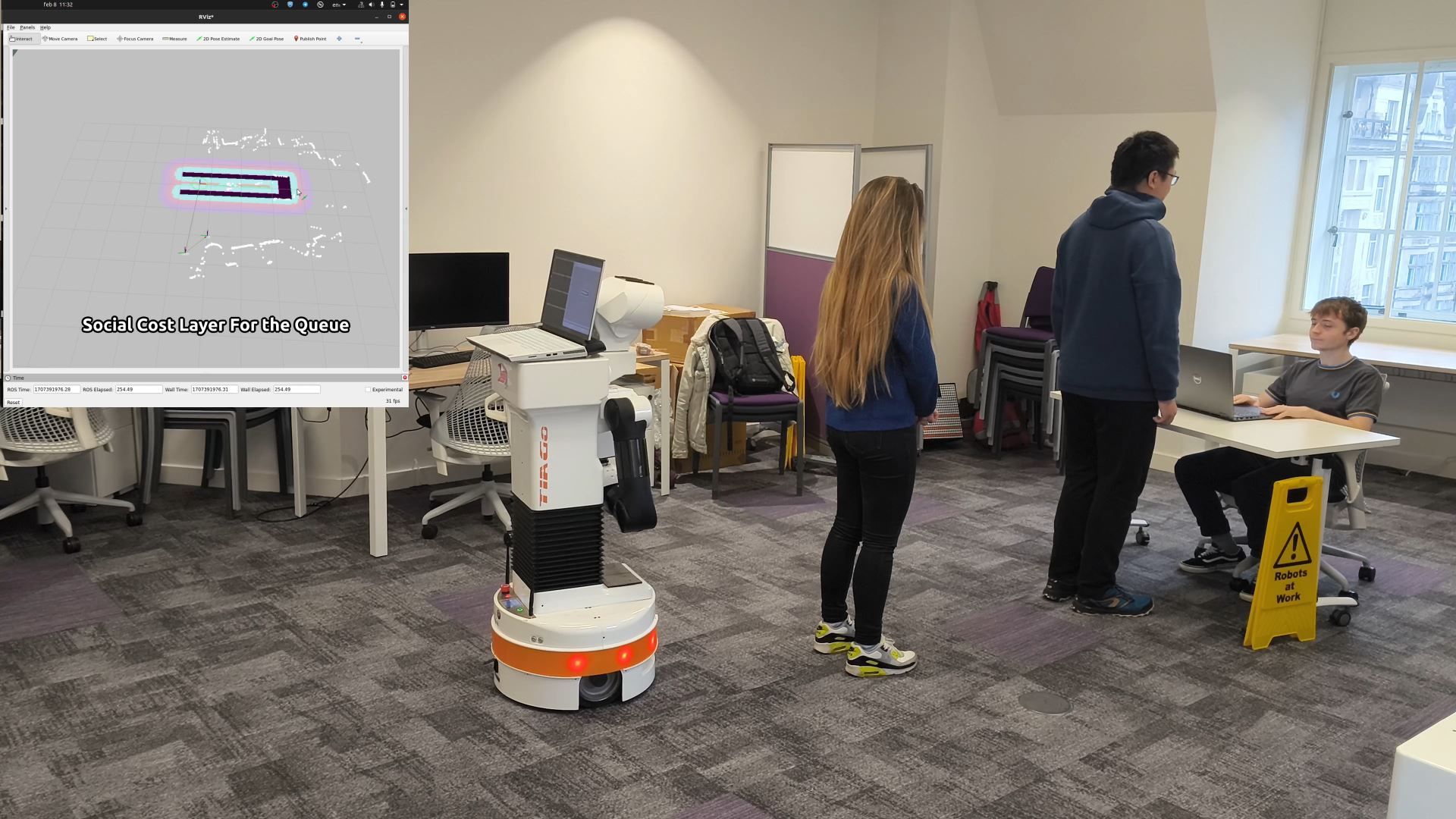} 
    \caption{Tests carried out on the robot in a real environment. With the social cost map enabled, the robot can approach the queue from the beginning, stay behind the last person as the line proceeds, and wait for its turn until the goal at the end of the queue is reached.}
    \label{fig:real_test}
\end{figure}

\subsection{Simulated Scenarios}
We now demonstrate the system in several practical scenarios in simulation, comparing the generated plan with and without the learned social costs. We built Gazebo \cite{koenig2004design} simulation environments containing a queue of people, a variable number of groups of people interacting, and other people passing by (Figure \ref{fig:general_sim}a). The robot mounts a lidar sensor to detect physical obstacles around it, while the information regarding the position of people is obtained directly from the simulator. 

In Figure \ref{fig:general_sim}, the robot is given a goal in front of a queue. The network can recognize and delimit the queue as well as the area between each group of people talking. Given the social obstacles, the planner computes a path around the groups and the line of people, queuing correctly behind the last person in line. In comparison, without the social layer, the robot plans between the people talking and directly to the goal, cutting the line (Figure \ref{fig:general_sim}c).

We further demonstrate the queue task by approaching the line of people from different angles. The robot is positioned in front, on the side, and behind different queues containing a variable number of people. As seen from Figure \ref{fig:approaching_direction}, the network can correctly recognize the queue regardless of where the robot comes from. Furthermore, the provided social cost map allows the planner to compute an appropriate path for the robot to queue correctly, position behind the last person, and follow the line of people until the goal is reached. Although the maximum number of people appearing in training set queues is 5, the network can correctly recognize and delimit a queue with any number of people, as long as the whole line is sufficiently aligned and contained within the input image (Figure \ref{fig:long_queue}$(b)$). 

We further demonstrate the talking groups scenario in  Figure \ref{fig:long_queue}$(a)$. The network can correctly identify different groups formed by a variable number of people. The network's virtual obstacles often have irregular shapes, but they are sufficient to prevent the robot from crossing the interaction zone.

A series of limitations of the network are dictated by the dataset used for its training. For example, the network cannot correctly recognize the queue if it is not settled along a straight line, since our dataset only contained straight queues. Our tests demonstrated that as long as each person's deviation from the center of the image does not exceed 1m (from both sides), the network can classify the queue correctly, but it often fails otherwise. Varying the maximum distance between each person in the queue could also lead to instability of the recognition. We noted that recognition starts failing when the mutual distance between people in the line is around 3m. Furthermore, while groups of people not in a queue but near one do not usually interfere, they may be interpreted as a second queue if placed near the goal. Similarly, if a goal is given too close to a particularly aligned group, it could be misinterpreted as a line of people.

Other incorrect results could arise in the presence of a crowd of people. The network can correctly recognize multiple groups of people and set the appropriate costs for each interaction space, even when such groups are close to each other. However, if the distances between people of different groups are smaller than the distances of people within a group, the network can produce incorrect virtual obstacles. Further refinement of the dataset can overcome some of these limitations, but if the environment is extremely crowded, other strategies that require the robot to ask people to move become necessary.

\subsection{Real-robot deployment}

The same system used in the simulation environment was deployed on a PAL Robotics TIAGo robot. Similarly to the simulation, the Lidar sensor detects obstacles around the robot. The information regarding the number and position of people in the scene are extracted from YoloV8 \cite{Jocher_Ultralytics_YOLO_2023}, using an RGB-D camera mounted on TIAGo's head. We recreated the tests made in simulation in the real environment (one example is shown in Figure \ref{fig:real_test}). 

For the queue task, once the people in line were correctly recognized and localized, the robot was given different starting points: in front, on the side, and behind the queue. Also in the real-life setting, the network recognized and delimited the queue correctly. The robot entered the line of people and followed it until the goal was reached, independently from the direction it came from.

We recreated a scenario with multiple groups with different numbers of people. A goal was provided at the end of the map. The network delimited the interaction areas, and the robot reached the goal without crossing any group of people.

The reliability of the perception and tracking system may limit the correctness of the computed social cost. In the queuing task, when the robot enters the line of people, the last person in the queue can obstruct the vision of the people behind. For this reason, the social grid map cannot be just generated from perceptions each time. To mitigate this problem, we keep track of every person seen in the past time instants, similarly to how the navigation stack remembers obstacles that are not presently in view. We place a marker in the last position where a person was detected and remove it when the area is in view and the laser sees no obstacle.

\section{Conclusions} \label{sec:conclusions}

This paper proposes a novel method to enhance Social Robotic Navigation by modifying the planner's cost function. Differently from past literature, we do not focus on proxemics or estimates of people's movement based on kinematic information. Instead, we consider areas the robot should plan to avoid based on the position of people in the environment, in accordance to specific social norms. We designed a modular plug-in that can add this capability seamlessly to virtually any navigation system. We developed a proof of concept based on two common social tasks: queuing in a line of people and respecting interaction spaces of people talking with each other. We demonstrate the effectiveness of our method both in simulation and in a real-world scenario, comparing it with traditional goal-based navigation. We focus on learning and navigation, and use off-the-shelf libraries for computer vision. The social maps could be further refined if the vision module was able to detect more than people's positions, for instance, whether or not they were talking, and with whom. 

Adapting the robot's behavior to what people are doing in the environment has not received as much attention as how people are moving. Some social norms affect navigation despite people not moving at all. We expect this line of research to receive more attention within social navigation in the near future.

\section*{Acknowledgment}
The work presented in this paper has born from the collaboration between the PIC4SeR Centre for Service Robotics at Politecnico di Torino and the Sensible Robots Research Group at King's College London.

%Bibliography
\bibliographystyle{unsrt}  
\bibliography{references}  

\begin{thebibliography}{10}

\bibitem{koenig2004design}
Design and use paradigms for gazebo, an open-source multi-robot simulator.
\newblock In {\em 2004 IEEE/RSJ international conference on intelligent robots and systems (IROS)(IEEE Cat. No. 04CH37566)}, volume~3, pages 2149--2154. IEEE, 2004.

\bibitem{burgard1999museum}
Wolfram Burgard, Armin~B Cremers, Dieter Fox, Dirk H{\"a}hnel, Gerhard Lakemeyer, Dirk Schulz, Walter Steiner, and Sebastian Thrun.
\newblock The museum tour-guide robot rhino.
\newblock In {\em Autonome Mobile Systeme 1998: 14. Fachgespr{\"a}ch Karlsruhe, 30. November--1. Dezember 1998}, pages 245--254. Springer, 1999.

\bibitem{thrun2000probabilistic}
Sebastian Thrun, Michael Beetz, Maren Bennewitz, Wolfram Burgard, Armin~B Cremers, Frank Dellaert, Dieter Fox, Dirk Haehnel, Chuck Rosenberg, Nicholas Roy, et~al.
\newblock Probabilistic algorithms and the interactive museum tour-guide robot minerva.
\newblock {\em The international journal of robotics research}, 19(11):972--999, 2000.

\bibitem{kirby2010social}
Rachel Kirby.
\newblock {\em Social robot navigation}.
\newblock Carnegie Mellon University, 2010.

\bibitem{kollmitz2015time}
Marina Kollmitz, Kaijen Hsiao, Johannes Gaa, and Wolfram Burgard.
\newblock Time dependent planning on a layered social cost map for human-aware robot navigation.
\newblock In {\em 2015 European Conference on Mobile Robots (ECMR)}, pages 1--6. IEEE, 2015.

\bibitem{scandolo2011anthropomorphic}
Leonardo Scandolo and Thierry Fraichard.
\newblock An anthropomorphic navigation scheme for dynamic scenarios.
\newblock In {\em 2011 IEEE International Conference on Robotics and Automation}, pages 809--814. IEEE, 2011.

\bibitem{fang2020human}
Fang Fang, Manxiang Shi, Kun Qian, Bo~Zhou, and Yahui Gan.
\newblock A human-aware navigation method for social robot based on multi-layer cost map.
\newblock {\em International Journal of Intelligent Robotics and Applications}, 4:308--318, 2020.

\bibitem{mateus2019efficient}
Andre Mateus, David Ribeiro, Pedro Miraldo, and Jacinto~C Nascimento.
\newblock Efficient and robust pedestrian detection using deep learning for human-aware navigation.
\newblock {\em Robotics and Autonomous Systems}, 113:23--37, 2019.

\bibitem{truong2016dynamic}
Xuan-Tung Truong and Trung-Dung Ngo.
\newblock Dynamic social zone based mobile robot navigation for human comfortable safety in social environments.
\newblock {\em International Journal of Social Robotics}, 8:663--684, 2016.

\bibitem{helbing1995social}
Dirk Helbing and Peter Molnar.
\newblock Social force model for pedestrian dynamics.
\newblock {\em Physical review E}, 51(5):4282, 1995.

\bibitem{ferrer2014proactive}
Gonzalo Ferrer and Alberto Sanfeliu.
\newblock Proactive kinodynamic planning using the extended social force model and human motion prediction in urban environments.
\newblock In {\em 2014 IEEE/RSJ International Conference on Intelligent Robots and Systems}, pages 1730--1735. IEEE, 2014.

\bibitem{yang2019socially}
Chun-Tang Yang, Tianshi Zhang, Li-Pu Chen, and Li-Chen Fu.
\newblock Socially-aware navigation of omnidirectional mobile robot with extended social force model in multi-human environment.
\newblock In {\em 2019 IEEE International Conference on Systems, Man and Cybernetics (SMC)}, pages 1963--1968. IEEE, 2019.

\bibitem{gupta2018social}
Agrim Gupta, Justin Johnson, Li~Fei-Fei, Silvio Savarese, and Alexandre Alahi.
\newblock Social gan: Socially acceptable trajectories with generative adversarial networks.
\newblock In {\em Proceedings of the IEEE conference on computer vision and pattern recognition}, pages 2255--2264, 2018.

\bibitem{mohamed2020social}
Abduallah Mohamed, Kun Qian, Mohamed Elhoseiny, and Christian Claudel.
\newblock Social-stgcnn: A social spatio-temporal graph convolutional neural network for human trajectory prediction.
\newblock In {\em Proceedings of the IEEE/CVF conference on computer vision and pattern recognition}, pages 14424--14432, 2020.

\bibitem{zhao2020noticing}
Dapeng Zhao and Jean Oh.
\newblock Noticing motion patterns: A temporal cnn with a novel convolution operator for human trajectory prediction.
\newblock {\em IEEE Robotics and Automation Letters}, 6(2):628--634, 2020.

\bibitem{vemula2018social}
Anirudh Vemula, Katharina Muelling, and Jean Oh.
\newblock Social attention: Modeling attention in human crowds.
\newblock In {\em 2018 IEEE international Conference on Robotics and Automation (ICRA)}, pages 4601--4607. IEEE, 2018.

\bibitem{bai2015intention}
Haoyu Bai, Shaojun Cai, Nan Ye, David Hsu, and Wee~Sun Lee.
\newblock Intention-aware online pomdp planning for autonomous driving in a crowd.
\newblock In {\em 2015 ieee international conference on robotics and automation (icra)}, pages 454--460. IEEE, 2015.

\bibitem{park2016hi}
Chonhyon Park, Jan Ond{\v{r}}ej, Max Gilbert, Kyle Freeman, and Carol O'Sullivan.
\newblock Hi robot: Human intention-aware robot planning for safe and efficient navigation in crowds.
\newblock In {\em 2016 IEEE/RSJ International Conference on Intelligent Robots and Systems (IROS)}, pages 3320--3326. IEEE, 2016.

\bibitem{trautman2015robot}
Pete Trautman, Jeremy Ma, Richard~M Murray, and Andreas Krause.
\newblock Robot navigation in dense human crowds: Statistical models and experimental studies of human--robot cooperation.
\newblock {\em The International Journal of Robotics Research}, 34(3):335--356, 2015.

\bibitem{mavrogiannis2019multi}
Christoforos~I Mavrogiannis and Ross~A Knepper.
\newblock Multi-agent path topology in support of socially competent navigation planning.
\newblock {\em The International Journal of Robotics Research}, 38(2-3):338--356, 2019.

\bibitem{mavrogiannis2020multi}
Christoforos~I Mavrogiannis and Ross~A Knepper.
\newblock Multi-agent trajectory prediction and generation with topological invariants enforced by hamiltonian dynamics.
\newblock In {\em Algorithmic Foundations of Robotics XIII: Proceedings of the 13th Workshop on the Algorithmic Foundations of Robotics 13}, pages 744--761. Springer, 2020.

\bibitem{chen2017decentralized}
Yu~Fan Chen, Miao Liu, Michael Everett, and Jonathan~P How.
\newblock Decentralized non-communicating multiagent collision avoidance with deep reinforcement learning.
\newblock In {\em 2017 IEEE international conference on robotics and automation (ICRA)}, pages 285--292. IEEE, 2017.

\bibitem{everett2018motion}
Michael Everett, Yu~Fan Chen, and Jonathan~P How.
\newblock Motion planning among dynamic, decision-making agents with deep reinforcement learning.
\newblock In {\em 2018 IEEE/RSJ International Conference on Intelligent Robots and Systems (IROS)}, pages 3052--3059. IEEE, 2018.

\bibitem{tai2018socially}
Lei Tai, Jingwei Zhang, Ming Liu, and Wolfram Burgard.
\newblock Socially compliant navigation through raw depth inputs with generative adversarial imitation learning.
\newblock In {\em 2018 IEEE international conference on robotics and automation (ICRA)}, pages 1111--1117. IEEE, 2018.

\bibitem{chen2019crowd}
Changan Chen, Yuejiang Liu, Sven Kreiss, and Alexandre Alahi.
\newblock Crowd-robot interaction: Crowd-aware robot navigation with attention-based deep reinforcement learning.
\newblock In {\em 2019 international conference on robotics and automation (ICRA)}, pages 6015--6022. IEEE, 2019.

\bibitem{pmlr-v205-xiao23a}
Xuesu Xiao, Tingnan Zhang, Krzysztof~Marcin Choromanski, Tsang-Wei~Edward Lee, Anthony Francis, Jake Varley, Stephen Tu, Sumeet Singh, Peng Xu, Fei Xia, Sven~Mikael Persson, Dmitry Kalashnikov, Leila Takayama, Roy Frostig, Jie Tan, Carolina Parada, and Vikas Sindhwani.
\newblock Learning model predictive controllers with real-time attention for real-world navigation.
\newblock In Karen Liu, Dana Kulic, and Jeff Ichnowski, editors, {\em Proceedings of The 6th Conference on Robot Learning}, volume 205 of {\em Proceedings of Machine Learning Research}, pages 1708--1721. PMLR, 14--18 Dec 2023.

\bibitem{choromanski2020rethinking}
Krzysztof Choromanski, Valerii Likhosherstov, David Dohan, Xingyou Song, Andreea Gane, Tamas Sarlos, Peter Hawkins, Jared Davis, Afroz Mohiuddin, Lukasz Kaiser, et~al.
\newblock Rethinking attention with performers.
\newblock {\em arXiv preprint arXiv:2009.14794}, 2020.

\bibitem{nakauchi2002social}
Yasushi Nakauchi and Reid Simmons.
\newblock A social robot that stands in line.
\newblock {\em Autonomous Robots}, 12:313--324, 2002.

\bibitem{banisetty2021socially}
Santosh~Balajee Banisetty, Scott Forer, Logan Yliniemi, Monica Nicolescu, and David Feil-Seifer.
\newblock Socially aware navigation: A non-linear multi-objective optimization approach.
\newblock {\em ACM Transactions on Interactive Intelligent Systems (TiiS)}, 11(2):1--26, 2021.

\bibitem{9515424}
Santosh~Balajee Banisetty, Vineeth Rajamohan, Fausto Vega, and David Feil-Seifer.
\newblock A deep learning approach to multi-context socially-aware navigation.
\newblock In {\em 2021 30th IEEE International Conference on Robot \& Human Interactive Communication (RO-MAN)}, pages 23--30, 2021.

\bibitem{kendon1990conducting}
Adam Kendon.
\newblock {\em Conducting interaction: Patterns of behavior in focused encounters}, volume~7.
\newblock CUP Archive, 1990.

\bibitem{cristani2011social}
Marco Cristani, Loris Bazzani, Giulia Paggetti, Andrea Fossati, Diego Tosato, Alessio Del~Bue, Gloria Menegaz, and Vittorio Murino.
\newblock Social interaction discovery by statistical analysis of f-formations.
\newblock In {\em BMVC}, volume~2, pages 10--5244, 2011.

\bibitem{setti2015f}
Francesco Setti, Chris Russell, Chiara Bassetti, and Marco Cristani.
\newblock F-formation detection: Individuating free-standing conversational groups in images.
\newblock {\em PloS one}, 10(5):e0123783, 2015.

\bibitem{hedayati2020reform}
Hooman Hedayati, Annika Muehlbradt, Daniel~J Szafir, and Sean Andrist.
\newblock Reform: Recognizing f-formations for social robots.
\newblock In {\em 2020 IEEE/RSJ International Conference on Intelligent Robots and Systems (IROS)}, pages 11181--11188. IEEE, 2020.

\bibitem{hedayati2019recognizing}
Hooman Hedayati, Daniel Szafir, and Sean Andrist.
\newblock Recognizing f-formations in the open world.
\newblock In {\em 2019 14th ACM/IEEE International Conference on Human-Robot Interaction (HRI)}, pages 558--559. IEEE, 2019.

\bibitem{taylor2020robot}
Angelique Taylor, Darren~M Chan, and Laurel~D Riek.
\newblock Robot-centric perception of human groups.
\newblock {\em ACM Transactions on Human-Robot Interaction (THRI)}, 9(3):1--21, 2020.

\bibitem{schmuck2020rica}
Viktor Schmuck and Oya Celiktutan.
\newblock Rica: Robocentric indoor crowd analysis dataset.
\newblock {\em IMU}, 127(74,234):31--172, 2020.

\bibitem{schmuck2021growl}
Viktor Schmuck and Oya Celiktutan.
\newblock Growl: Group detection with link prediction.
\newblock In {\em 2021 16th IEEE International Conference on Automatic Face and Gesture Recognition (FG 2021)}, pages 1--8. IEEE, 2021.

\bibitem{schmuck2022igrowl}
Viktor Schmuck and Oya Celiktutan.
\newblock igrowl: Improved group detection with link prediction.
\newblock {\em IEEE Transactions on Biometrics, Behavior, and Identity Science}, 2022.

\bibitem{ros2}
Steven Macenski, Tully Foote, Brian Gerkey, Chris Lalancette, and William Woodall.
\newblock Robot operating system 2: Design, architecture, and uses in the wild.
\newblock {\em Science Robotics}, 7(66):eabm6074, 2022.

\bibitem{macenski2020marathon2}
Steven Macenski, Francisco Martin, Ruffin White, and Jonatan Ginés~Clavero.
\newblock The marathon 2: A navigation system.
\newblock In {\em 2020 IEEE/RSJ International Conference on Intelligent Robots and Systems (IROS)}, 2020.

\bibitem{Jocher_Ultralytics_YOLO_2023}
Glenn Jocher, Ayush Chaurasia, and Jing Qiu.
\newblock {Ultralytics YOLO}, January 2023.

\end{thebibliography}

\end{document}